\pdfoutput=1

\documentclass[11pt]{article}

\usepackage{acl}

\usepackage{times}
\usepackage{latexsym}

\usepackage[T1]{fontenc}

\usepackage[utf8]{inputenc}

\usepackage{microtype}

\usepackage{booktabs}
\usepackage{graphicx}

\RequirePackage{amsmath,amsfonts,amssymb}

\title{SATLab at SemEval-2022 Task 4: Trying to Detect Patronizing and Condescending Language with only Character and Word N-grams}

\author{Yves Bestgen \\
  Statistical Analysis of Text Laboratory (SATLab) \\
  Universit\'e catholique de Louvain \\
  Place Cardinal Mercier, 10 1348 Louvain-la-Neuve, Belgium \\
  \texttt{yves.bestgen@uclouvain.be} \\}

\begin{document}
\maketitle
\begin{abstract}
A logistic regression model only fed with character and word n-grams is proposed for the SemEval-2022 Task 4 on Patronizing and Condescending Language Detection (PCL). It obtained an average level of performance, well above the performance of a system that tries to guess without using any knowledge about the task, but much lower than the best teams. As the proposed model is very similar to the one that performed well on a task requiring to automatically identify hate speech and offensive content, this paper confirms the difficulty of PCL detection.

\end{abstract}

\section{Introduction}

This paper presents the SATLab's participation in SemEval-2022 Task 4: Patronizing and Condescending Language Detection. It is a very recent task that aims at identifying passages in texts in which a person is condescending to a vulnerable community \citep{perezalmendros2020dont,wang2019talkdown}. The task organizers have proposed to identify this type of discourse in paragraphs extracted from news stories published in English-speaking media.

The Patronizing and Condescending Language (PCL) Detection task is part of the large family of tasks that aim at automatically identifying problematic language, whether in media or on social networks. Examples include the detection of hate speech and offensive content, fakenews and hyperpartisan news \citep{DBLP:conf/fire/0001MMPDMP19,kiesel2019}. The PCL challenge, however, differs from these cases in one important characteristic. As pointed out by \citep{perezalmendros2020dont}, using PCL is not always conscious and the intention of its author is often positive. Nevertheless, PCL tends to indirectly demean the community in question and reinforce negative stereotypes about it. It is therefore important to develop procedures to identify it.

Recently, the SATLab performed well on a task requiring to automatically identify hate speech and offensive content in tweets using a classical supervised algorithm only fed with character n-grams \citep{BYFIRE21}. It was especially effective for resource-poor languages (e.g., Marathi). For English, its performance ($F1 = 0.782$) was average, but nevertheless relatively close to the best team ($F1 = 0.831$) that used complementary resources, pre-trained embeddings and deep learning approaches. These performances suggest that the proposed approach could be a interesting baseline to evaluate the benefits of these more complex approaches. 

The goal of SATLab's participation in the PCL detection task was to evaluate whether this finding can be generalized to this new task. A priori, one might think that the performance of the system should be significantly worse here for two reasons. First, this task seems to be much more difficult because of the complex language means of expression on which it relies and the considerable amount of world knowledge and common sense reasoning required to understand this type of language \citep{perezalmendros2020dont}. Secondly, this task is only proposed for English, a language for which many additional resources are available.
 
The rest of this document presents the task (see \citet{perezalmendros2022semeval} for more details), the proposed system and the performances reached in the challenge.

\section{Task} 

For this challenge, the organizers extracted paragraphs mentioning predefined vulnerable communities, such as \textit{homeless people}, \textit{migrants} or \textit{poor families}, in news stories from various media in different countries, all in English. Each paragraph was annotated according to whether or not it contained one or more instances of PCL. Annotators were also asked to identify the type of PCLs present in a paragraph using seven categories.

On this materials, the SATLab participated in the two tasks proposed by the organizers. The first task consists in deciding whether a paragraph contains some kind of PCL or not. It is thus a classical binary decision task. 

The second task is a fine categorization task in seven positive and one negative categories. The seven positive PCL categories are: \textit{a)}\footnote{The letters are used to identify these categories in Table 2 and 4.} \textit{Unbalanced power relations, b) Shallow solution, c) Presupposition, d) Authority voice, e) Metaphor, f) Compassion, g) The poorer, the merrier}. It is important to note that this categorization is non-exclusive: a single instance can be an example of two, three, four and even five categories. I treated this second problem as Task 1 and thus as seven independent binary problems, the predictions of the seven models being simply concatenated in the final submission. For this second task, the organizers also provided the precise position of the text areas that had been identified by the annotators as warranting the assignment of a given category. This information was not used.

The dataset provided by the organizers to develop the systems consisted of 10,469 instances (but one, $id =	@@16852855$, contained no text). It was highly unbalanced for Task 1 with only 9.5\% positive instances. The distribution was even more unbalanced in Task 2 since it contained the same proportion of negative cases. The most frequent positive category (a) represented 6.84\% of the total instances and the least frequent only 0.38\%. 

For the system development phase, the organizers proposed a division of the paragraphs into a learning set (80\%) and a development set. The test set used in the final evaluation of the systems, whose responses were therefore unknown, consisted of 3,832 paragraphs.

The measure of effectiveness chosen by the organizers was the F1-score on the positive category for Task 1 and the unweighted average of the F1-scores on the seven PCL categories for Task 2.

\section{System}
The proposed system is adapted from the SATLab's participation at HASOC 2021. It is based on the following components.

\begin{table}
\centering
\caption{Performance on the development set}
\label{tab:par1}
\begin{tabular}{cccc}
\toprule
Condition & Char & Word & Combi \\
\midrule
F1 & 0.443 & 0.440 & 0.468 \\
\bottomrule
\end{tabular}
\end{table} 

\subsection{Features}
The only features used were character and word n-grams. These n-grams were extracted from the lowercased paragraphs. The character n-grams had a length between 1 and 7. The extracted word n-grams contained from 1 to 4 words. The tokenization provided in the materials was used. All n-grams present at least twice in the materials were extracted. 
 
\subsection{Weighting schema}
BM25 (for Best Match 25) was used to weight the frequency of the features in each paragraph \citep{Robertson:2009:PRF,bestgen:2017:VarDial}. It is a kind of TF-IDF with specific choices for each of the two components, but above all it takes into account the length of the document. Its classical formula is:
\begin{equation}
\begin{split}
   \text{BM25} = \frac{tf}{tf + k_1 * (1 - b + b * \frac{dl}{dl-avg_{dl}})} \\ 
   \times\log\frac{N - df + 0.5}{df + 0.5} 
\end{split}
\end{equation}
in which $tf$ refers to the frequency of the term in the paragraph, $N$ is the number of paragraphs in the set, $df$ the number of paragraphs that include the term, $dl$ the length of the paragraph and $avg_{dl}$ the average length of the paragraphs in the set. The parameter $k_1$ was set to 2 and $b$ to 0.75.
 
\subsection{Regularization}
 The feature values for each paragraph were regularized using the L2 norm.
\begin{table*}
\centering
\caption{Parameters for the two submissions for each task}
\label{tab:par1}
\begin{tabular}{llrrrrrrrr}
\toprule
Sub & P & T1 & T2a & T2b & T2c & T2d & T2e & T2f & T2g \\ 
\midrule
1 & C & 3.1 & 4.75 & 0.95 & 0.55 & 0.35 & 0.25 & 0.95 & 0.014\\
& w1 & 180 & 500 & 1,600 & 1,300 & 700 & 1,250 & 850 & 1,400\\
\midrule
2 & C & 2 & 3.75 & 0.90 & 0.70 & 0.65 & 0.40 & 1.45 & 0.016\\
& w1 & 50 & 300 & 2,000 & 1,500 & 1,400 & 750 & 1,750 & 1,800\\
\bottomrule
\end{tabular}
\end{table*} 
 
\subsection{Supervised learning procedure}
These character and word n-grams were the only features provided to the supervised learning procedure: the L2-regularized logistic regression as implemented in the LIBLinear package \citep{Fan2008}. This procedure is extremely fast and very simple to implement because it only requires the optimization of two parameters: the regularization parameter $C$ and $-wi$ which allows to adjust this parameter $C$ for the positive category, the one which has the most influence in the efficiency measure. It should be noted that during the development period of the system, tests were carried out with an approach much slower and much more complex to optimize: a gradient boosting decision tree as implemented in the LightGBM free software \citep{LightGBM}. But, this approach was abandoned because it did not improve the efficiency of the system.

\subsection{Comparison to HASOC}
The main difference between this system and the one used for HASOC 2021 is the addition of word n-gram. This decision was made during the development phase of the system, carried out on the basis of the division of the materials into a train set and development set as provided by the organizers. Table 1 shows the F1-scores on the positive category for the models based on each n-gram type and their combination. The parameters were optimized directly on the development set, which obviously raises the concern of overfit, but it can be assumed to be similar for each condition compared. As we can see, the two types of n-grams produce very similar performances and the combination brings a small benefit. The addition of word n-grams means that the system can no longer be considered completely language agnostic like the HASOC system, because it relies on the tokenization provided by the organizers and because a number of characters, such as punctuation marks, are removed.

\begin{table}
\centering
\caption{Results for Task 1 (N = 79)}
\label{tab:res1}
\begin{tabular}{lllll}
\toprule
Rank & Id  & Prec. & Rec. & F1-Score \\
\midrule
1 & First & 0.646 & 0.656 & 0.651 \\
43 & Baseline & 0.394 & 0.653 & 0.491\\
54 & SATLab & 0.348 & 0.552 & 0.427\\
\bottomrule
\end{tabular}
\end{table} 

\begin{table*}
\centering
\caption{Results for Task 2 (N = 49)}
\label{tab:res1}
\begin{tabular}{llllllllll}
\toprule
Rank & Id & a & b & c & d & e & f & g & Mean F1 \\
\midrule
1 & First & 0.656 & 0.529 & 0.369 & 0.407 & 0.359 & 0.492 & 0.471 & 0.469\\
24 & SATLab & 0.424 & 0.331 & 0.170 & 0.232 & 0.175 & 0.315 & 0.142 & 0.256\\
38 & Baseline & 0.354 & 0 & 0.167 & 0 & 0 & 0.209 & 0 & 0.104\\
\bottomrule
\end{tabular}
\end{table*}

\section{Results}

The system just described is identical for the two tasks, the only differences being in the parameters of the logistic regression which were optimized independently for each task and for each target category in Task 2. For the two final submissions, these parameters were optimized by a 5-fold cross-validation procedure applied to the combination of the training and development sets using several steps of exhaustive grid search.

The parameters employed for each submission are given in Table 2. The parameters for the first submission were those that produced the best overall performance in the cross-validation experiments while those for the second submission corresponded to a model producing close performance, but in which the difference between precision and recall for the target categories was as small as possible.

As can be seen in this table, the parameter values are often very different in the two versions while the performance of the models was very close.

Tables 3 and 4 show the performance of the better of the two submissions on the test set, as provided by the organizers. For Task 1, submission 1 performed better while submission 2 performed better for Task 2. The differences between the two submissions for the two tasks are however very small, less than 0.005. These two tables also give the performance of the first team in the challenge as well as that of the Roberta-based Baseline proposed by the organizers.

The proposed system outperformed the Roberta-based Baseline in Task 2, but not in Task 1. It ranked in the middle of the participants at best, very far from the challenge winners. It is unfortunate (for the SATLab at least) that the task was not proposed in languages with less resources and precomputed embeddings, making the use of Deep Learning much more easier.

It is interesting to note that the performances obtained on the test set is only slightly lower than those obtained when optimizing the parameters by a cross-validation procedure on the training set. For task 1, this performance on the training set was only 0.03 higher and, on task 2, only 0.008. This suggests that the optimization did not produce an overfit. 

The comparison of the system's performance to that of the best team for the seven categories (Table 4) shows that for none of the categories does the system manage to come close to this benchmark. The differences are always approximately 0.20, except \textit{a) Unbalanced power relations}, which is by far the most frequent category, and for \textit{g) The poorer, the merrier}, which is by far the rarest in the learning set, categories for which they are even higher. The more than limited effectiveness of the system does not seem to justify a detailed analysis of its errors.

While the F1-score has become the standard of evaluation for NLP categorization tasks, its interpretation is not obvious. One may wonder whether a Mean F1 of 0.256 (Task 2) represents a prediction of at least an acceptable value. A priori, this does not seem to be the case. On the other hand, the large differences in the frequency of the categories and the imbalance in favor of the negative category may explain the relatively low F1. One way to answer this question more objectively is to determine the best level of performance a system that tries to guess without using any knowledge about the task (Best Guess) can expect to achieve. Ansgar Grüne in his blog post available at \url{https://inside.getyourguide.com/blog/2020/9/30/what-makes-a-good-f1-score} shows\footnote{I confirmed empirically this analysis on the task materials by means of a Monte-Carlo procedure in which the proportion of instances in the positive categories were varied between 0 and 1.} that in the case of a binary task this is the F1 obtained by a system that always predicts the positive category. If $q$ is the actual proportion of instances belonging to this positive category, 
\begin{equation}
   \text{F1 Best Guess} = \frac{2q}{q + 1}. 
\end{equation}
This value can also be obtained by submitting the responses of a system that always predicts the positive category, the approach I used here since the challenge participants still ignore the proportion of each category in the test set. 

In the present challenge, this F1 Best Guess is 0.153 for Task 1 and 0.041 for Task 2. A system that would not exceed these values therefore does no better than a system that always predicts the positive category. SATLab's performance is five times better than this baseline for Task 2, but less than three times better for Task 1. In absolute value, the gain compared to the Best Guess is 0.274 for Task 1 and 0.205 for Task 2. So the n-grams bring some information, but it is obvious that it is not enough to perform well in this task.

\section{Conclusion}
The SATLab's proposed system for SemEval-2022 Task 4: Patronizing and Condescending Language Detection relies solely on the character and word n-grams present in the paragraphs to be categorized. It does not use any additional data and employs a classical supervised learning procedure (i.e., logistic regression). It obtained an average level of performance, well above the performance of a system that tries to guess without using any knowledge about the task, but much lower than the best teams. Compared to the performance obtained in the HASOC task, it is clearly further from these best teams. These results confirm, if it were necessary, the much greater difficulty of PCL detection compared to hate speech and offensive content identification \citep{perezalmendros2020dont}. They thus suggest, unless other teams using a similar approach were more successful, that the use of much more complex approaches is essential for the PCL task.

\section*{Acknowledgements}
The author is a Research Associate of the Fonds de la Recherche Scientifique - FNRS (F\'ed\'eration Wallonie Bruxelles de Belgique).
\bibliography{custom}

\begin{thebibliography}{10}
\expandafter\ifx\csname natexlab\endcsname\relax\def\natexlab#1{#1}\fi

\bibitem[{Bestgen(2017)}]{bestgen:2017:VarDial}
Yves Bestgen. 2017.
\newblock Improving the character ngram model for the {DSL} task with {BM25}
  weighting and less frequently used feature sets.
\newblock In \emph{Proceedings of the Fourth Workshop on NLP for Similar
  Languages, Varieties and Dialects (VarDial)}, pages 115--123, Valencia,
  Spain.

\bibitem[{Bestgen(2021)}]{BYFIRE21}
Yves Bestgen. 2021.
\newblock A simple language-agnostic yet strong baseline system for hate speech
  and offensive content identification.
\newblock In \emph{Working Notes of {FIRE} 2021 - Forum for Information
  Retrieval Evaluation}, {CEUR} Workshop Proceedings. CEUR-WS.org.

\bibitem[{Fan et~al.(2008)Fan, Chang, Hsieh, Wang, and Lin}]{Fan2008}
Rong-En Fan, Kai-Wei Chang, Cho-Jui Hsieh, Xiang-Rui Wang, and Chih-Jen Lin.
  2008.
\newblock {LIBLINEAR}: A library for large linear classification.
\newblock \emph{Journal of Machine Learning Research}, 9:1871--1874.

\bibitem[{Ke et~al.(2017)Ke, Meng, Finley, Wang, Chen, Ma, Ye, and
  Liu}]{LightGBM}
Guolin Ke, Qi~Meng, Thomas Finley, Taifeng Wang, Wei Chen, Weidong Ma, Qiwei
  Ye, and Tie-Yan Liu. 2017.
\newblock \href
  {http://papers.nips.cc/paper/6907-lightgbm-a-highly-efficient-gradient-boosting-decision-tree.pdf}
  {{LightGBM}: A highly efficient gradient boosting decision tree}.
\newblock In I.~Guyon, U.~V. Luxburg, S.~Bengio, H.~Wallach, R.~Fergus,
  S.~Vishwanathan, and R.~Garnett, editors, \emph{Advances in Neural
  Information Processing Systems 30}, pages 3146--3154. Curran Associates, Inc.

\bibitem[{Kiesel et~al.(2019)Kiesel, Mestre, Shukla, Vincent, Adineh, Corney,
  Stein, and Potthast}]{kiesel2019}
Johannes Kiesel, Maria Mestre, Rishabh Shukla, Emmanuel Vincent, Payam Adineh,
  David Corney, Benno Stein, and Martin Potthast. 2019.
\newblock {SemEval-2019 Task 4: Hyperpartisan News Detection}.
\newblock In \emph{Proceedings of The 13th International Workshop on Semantic
  Evaluation {(SemEval 2019)}}. Association for Computational Linguistics.

\bibitem[{Mandl et~al.(2019)Mandl, Modha, Majumder, Patel, Dave, Mandalia, and
  Patel}]{DBLP:conf/fire/0001MMPDMP19}
Thomas Mandl, Sandip Modha, Prasenjit Majumder, Daksh Patel, Mohana Dave,
  Chintak Mandalia, and Aditya Patel. 2019.
\newblock \href {https://doi.org/10.1145/3368567.3368584} {Overview of the
  {HASOC} track at {FIRE} 2019: Hate speech and offensive content
  identification in indo-european languages}.
\newblock In \emph{{FIRE} '19: Forum for Information Retrieval Evaluation,
  Kolkata, India, December, 2019}, pages 14--17. {ACM}.

\bibitem[{P{\'e}rez-Almendros et~al.(2020)P{\'e}rez-Almendros, Espinosa-Anke,
  and Schockaert}]{perezalmendros2020dont}
Carla P{\'e}rez-Almendros, Luis Espinosa-Anke, and Steven Schockaert. 2020.
\newblock {Don’t Patronize Me! An Annotated Dataset with Patronizing and
  Condescending Language towards Vulnerable Communities}.
\newblock In \emph{Proceedings of the 28th International Conference on
  Computational Linguistics}, pages 5891--5902.

\bibitem[{P{\'e}rez-Almendros et~al.(2022)P{\'e}rez-Almendros, Espinosa-Anke,
  and Schockaert}]{perezalmendros2022semeval}
Carla P{\'e}rez-Almendros, Luis Espinosa-Anke, and Steven Schockaert. 2022.
\newblock {SemEval-2022 Task 4: Patronizing and Condescending Language
  Detection}.
\newblock In \emph{Proceedings of the 16th International Workshop on Semantic
  Evaluation (SemEval-2022)}. Association for Computational Linguistics.

\bibitem[{Robertson and Zaragoza(2009)}]{Robertson:2009:PRF}
Stephen Robertson and Hugo Zaragoza. 2009.
\newblock The probabilistic relevance framework: {BM25} and beyond.
\newblock \emph{Foundations and Trends in Information Retrieval},
  3(4):333--389.

\bibitem[{Wang and Potts(2019)}]{wang2019talkdown}
Zijian Wang and Christopher Potts. 2019.
\newblock \href {http://arxiv.org/abs/1909.11272} {Talkdown: A corpus for
  condescension detection in context}.

\end{thebibliography}

\end{document}